\documentclass[lettersize,journal]{IEEEtran}
\usepackage{amsmath,amsfonts}
\usepackage{algorithmic}
\usepackage{algorithm}
\usepackage{array}
\usepackage[caption=false,font=normalsize,labelfont=sf,textfont=sf]{subfig}
\usepackage{textcomp}
\usepackage{stfloats}
\usepackage{url}
\usepackage{verbatim}
\usepackage{graphicx}
\usepackage{cite}
\usepackage{hyperref}
\usepackage{booktabs}
\usepackage{caption}
\begin{document}

\title{Focaler-IoU: More Focused Intersection over Union Loss}

\author{Hao Zhang, Shuaijie Zhang}

\maketitle

\begin{abstract}
Bounding box regression plays a crucial role in the field of object detection, and the positioning accuracy of object detection largely depends on the loss function of bounding box regression. Existing researchs improve regression performance by utilizing the geometric relationship between bounding boxes, while ignoring the impact of difficult and easy sample distribution on bounding box regression. In this article, we analyzed the impact of difficult and easy sample distribution on regression results, and then proposed Focaler-IoU, which can improve detector performance in different detection tasks by focusing on different regression samples. Finally, comparative experiments were conducted using existing advanced detectors and regression methods for different detection tasks, and the detection performance was further improved by using the method proposed in this paper.Code is available at \url{https://github.com/malagoutou/Focaler-IoU}.

\end{abstract}


\section{Introduction}
\IEEEPARstart{O}{bject} detection is one of the basic tasks of computer vision, which aims to locate and recognize the object in an image. 
They can be categorized into Anchor-based and Anchor-free methods according to whether they generate an Anchor or not.
Anchor based algorithms include FasterR-CNN\cite{ref1}, YOLO (You Only Look Once) series\cite{ref2},SSD (Single Shot MultiBox Detector)\cite{ref3} and RetinaNet\cite{ref4}.Anchor free detection algorithms include CornerNet\cite{ref5},CenterNet\cite{ref6} and FCOS (Fully Convolutional One StageObject Detection)\cite{ref7}. In these detectors, the bounding box regression loss function plays an irreplaceable role as an important component of the localization branch.
\subsection{Bounding Box Regression Losses}
 With the development of computer vision, the object detection task has received more attention from researchers. In order to evaluate the performance of various algorithms on the detection task, a well-suited metric needs to be introduced. Before IoU (Intersection over Union)\cite{ref8} was proposed, $\textit{l}_n$-norm loss\cite{ref13} was used as the evaluation metric for early bounding box regression problems, however, since $\textit{l}_n$-norm loss is very sensitive to outliers, this leads to a greater impact of outliers on the loss, making the performance unstable when outliers exist in the model. To better solve the above problems, a more appropriate metric is proposed: IoU (Intersection over Union)\cite{ref8}. Under the evaluation criterion based on IoU, the detection accuracy of most object detection tasks is further improved, but there are some shortcomings in the IoU loss itself, for example, when there is no overlap between the GT box and the anchor, their gradient will disappear, which cannot accurately characterize the positional relationship between the two bounding boxes. To remedy this deficiency, GIoU\cite{ref9} proposes that using the minimum enclosing box containing the GT box and the anchor to calculate the loss can improve the detection performance. In CIoU and DIoU\cite{ref10}, in order to compensate for the slow convergence speed of GIoU, CIoU accelerates the convergence by further considering the aspect ratio between GT box and anchor, while DIoU accelerates the convergence by normalizing the distance between the centroids of two bounding boxes. EIoU\cite{ref12} further considers the shape loss on the basis of CIoU, which can accelerate the convergence by minimizing the difference between the width and height of GT box and anchor, and SIoU\cite{ref13} further considers the angle of the line connecting the centers of the two bounding boxes and redefines the distance loss and shape loss according to the angle, and adds them into the loss function as a new loss term, which achieves the best detection effect among the current IoU-based loss functions.
\subsection{Focal Loss}
The problem of unbalanced training samples persists in the process of marginal regression. The training samples can be categorized into positive and negative samples according to whether they contain the target category or not. Some conventional solutions for training sample imbalance are by sampling and reweighting difficult samples during the training process, however, the effect of this method is not significant. In Focal Loss\cite{ref14}, it is proposed that the easily recognized negative samples account for the majority of the total loss and dominate the gradient. Focal Loss\cite{ref14} improves the model's ability to recognize rare target categories by adjusting the weights of the positive and negative samples so that the model focuses more on the difficult-to-classify positive samples and reduces the weights of the negative samples that are relatively easy to classify. In Libra R-CNN\cite{ref15}, a simple and effective framework for balanced learning is proposed, where Balanced L1 loss is used at the objective level to classify the training samples into outliers and inliners. outliers are regarded as difficult samples, which can generate larger gradients compared to inliners, which is harmful to the training process. process. So Libra R-CNN uses gradient regression that promotes inliners and clips the large gradients generated by those outliers to achieve better classification results. In EIoU\cite{ref12}, the training samples are divided into high-quality samples (anchors) and low-quality samples (outliers), and FocalL1 loss is proposed on the basis of L1 loss to increase the gradient contribution of high-quality samples to the training process. Meanwhile, the EIoU loss is added as a variable to the FocalL1 loss, so that the model can pay more attention to high-quality samples to further improve the detection effect.
\par The main contributions of this article are as follows:
\par$\bullet$We analyzed the impact of the distribution of difficult and easy samples on bounding box regression. Based on existing bounding box regression methods, Focaler-IoU is proposed to focus on different regression samples through linear interval mapping.
\par$\bullet$We conducted experiments using advanced one-stage detectors to verify that our method can effectively improve detection performance and compensate for the shortcomings of existing methods.
\section{Related work}
In recent years, with the development of detectors, the edge regression loss has been rapidly developed. At first, IoU\cite{ref8} was proposed to be used for evaluating the border regression state, and then new constraints GIoU\cite{ref9}, DIoU\cite{ref10}, CIoU\cite{ref10}, EIoU\cite{ref12}, and SIoU\cite{ref11} were continuously added on the basis of IoU, etc. were successively proposed.
\subsection{IoU Metric}
\par The IoU (Intersection over Union)\cite{ref8}, which is the most popular target detection evaluation criterion, is defined as follows:
\begin{equation} 
	IoU=\displaystyle\frac{\left\vert B\cap B^{gt} \right\vert}{\left\vert B\cup B^{gt} \right\vert}
\end{equation}
where $B$ and $B^{gt}$  represent the predicted box and the GT box, respectively.
\subsection{GIoU Metric}
In order to solve the gradient vanishing problem of IoU loss due to no overlap between GT box and Anchor box in the bounding box regression, GIoU (Generalized Intersection over Union)\cite{ref9} is proposed. Its definition is as follows:
	\begin{equation} 
	GIoU=IoU-\displaystyle\frac{\left\vert C-B\cap B^{gt} \right\vert}{\left\vert C \right\vert}
\end{equation}
 where $C$ represents the smallest enclosing box between the GT box and the Anchor box.
 \subsection{DIoU Metric}
 Compared with GIoU, DIoU\cite{ref10} considers the distance constraints between the bounding boxes, and by adding the centroid normalized distance loss term on the basis of IoU, its regression results are more accurate. It is defined as follows:
 	\begin{equation}
 	DIoU=IoU-\displaystyle\frac{\rho^{2}(b,b^{gt})}{c^{2}}
 \end{equation}
 Where $b$ and $b^{gt}$  are the center points of anchor box and GT box respectively, $\rho\left(\cdot\right) $  refers to the Euclidean distance, where $c$ is the diagonal distance of the minimum enclosing bounding box between $b$ and $b^{gt}$.
 \par CIoU\cite{ref10} further considers the shape similarity between GT and Anchor boxes by adding a new shape loss term to DIoU to reduce the difference in aspect ratio between Anchor and GT boxes. It is defined as follows:
 \begin{equation}
 	CIoU=IoU-\displaystyle\frac{\rho^{2}(b,b^{gt})}{c^{2}}-\alpha v
 \end{equation}
 	\begin{equation}
 	\alpha=\displaystyle\frac{v}{(1-IoU)+v}
 \end{equation}
 \begin{equation}
 	v=\displaystyle\frac{4}{\pi^{2}}(arctan\displaystyle\frac{w^{gt}}{h^{gt}}-arctan\displaystyle\frac{w}{h})^{2}
 \end{equation}
 	where $w^{gt}$ and $h^{gt}$ denote the width and height of GT box,$w$ and $h$ denote the width and height of anchor box.
  \subsection{EIoU Metric}
  EIoU\cite{ref12} redefines the shape loss based on CIoU, and further improves the detection accuracy by directly reducing the aspect difference between GT boxes and Anchor boxes. It is defined as follows:
  \begin{equation}
  	EIoU=IoU-\displaystyle\frac{\rho^{2}(b,b^{gt})}{c^{2}}-\displaystyle\frac{\rho^{2}(w,w^{gt})}{(w^{c})^{2}}-\displaystyle\frac{\rho^{2}(h,h^{gt})}{(h^{c})^{2}}
  \end{equation}
   Where $w^{c}$ and $h^{c}$ are the width and height of the minimum bounding box covering GT box and anchor box.
    \subsection{SIoU Metric}
    On the basis of previous research, SIoU\cite{ref11} further considers the influence of the angle between the bounding boxes on the bounding box regression, which aims to accelerate the convergence process by decreasing the angle between the anchor box and GT box which is the horizontal or vertical direction. Its definition is as follows:
    	\begin{equation}
    	SIoU = IoU - \displaystyle\frac{(\Delta+\Omega)}{2}
    \end{equation}
    \begin{equation}
    	\Lambda=sin(2sin^{-1}\displaystyle\frac{min(\left\vert x_{c}^{gt}-x_{c} \right\vert,\left\vert y_{c}^{gt}-y_{c} \right\vert)}{\sqrt{ (x_{c}^{gt}-x_{c})^{2}+(y_{c}^{gt}-y_{c})^{2}}+\in})
    \end{equation}
    	\begin{equation}
    	\Delta=\sum_{t=w,h}(1-e^{-\gamma\rho_{t}}),\gamma=2-\Lambda
    \end{equation}
    	\begin{equation}
    	\left\{
    	\begin{aligned}
    		\displaystyle\rho_{x} & = & (\frac{x_{c}-x_{c}^{gt}}{w^{c}})^{2} \\
    		\rho_{y} & = & (\frac{y_{c}-y_{c}^{gt}}{h^{c}})^{2} \\
    	\end{aligned}
    	\right.
    \end{equation}
    	\begin{equation}
    	\Omega=\sum_{t=w,h}(1-e^{-\omega_{t}})^{\theta},\theta=4
    \end{equation}
    	\begin{equation}
    	\left\{
    	\begin{aligned}
    		\displaystyle\omega_{w} & = & \frac{\left\vert w - w_{gt} \right\vert}{max(w,w_{gt})} \\
    		\displaystyle\omega_{h} & = & \frac{\left\vert h - h_{gt} \right\vert}{max(h,h_{gt})} \\
    	\end{aligned}
    	\right.
    \end{equation}
\section{Methods}
\subsection{Analysis}
	\begin{figure*}[!htbp]
	\centering
	\subfloat[]{
		\centering
		\includegraphics[width=3.5in]{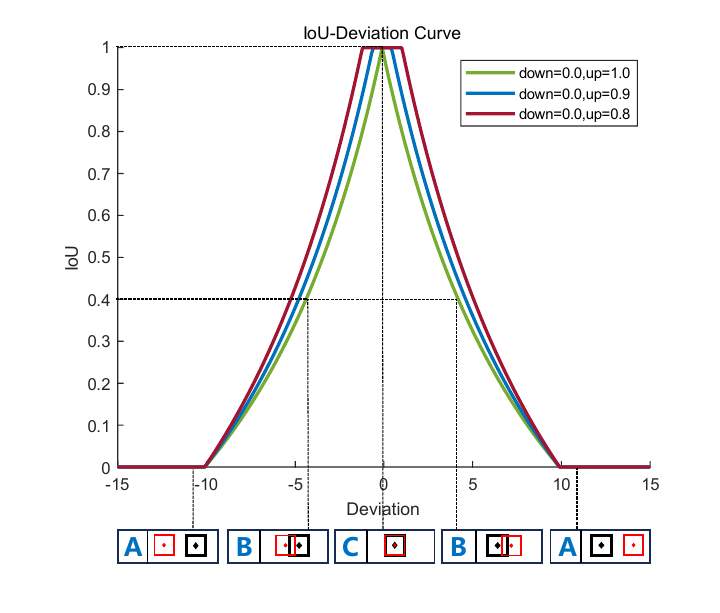}} 
		\hfill
	\subfloat[]{	
		\centering
		\includegraphics[width=3.5in]{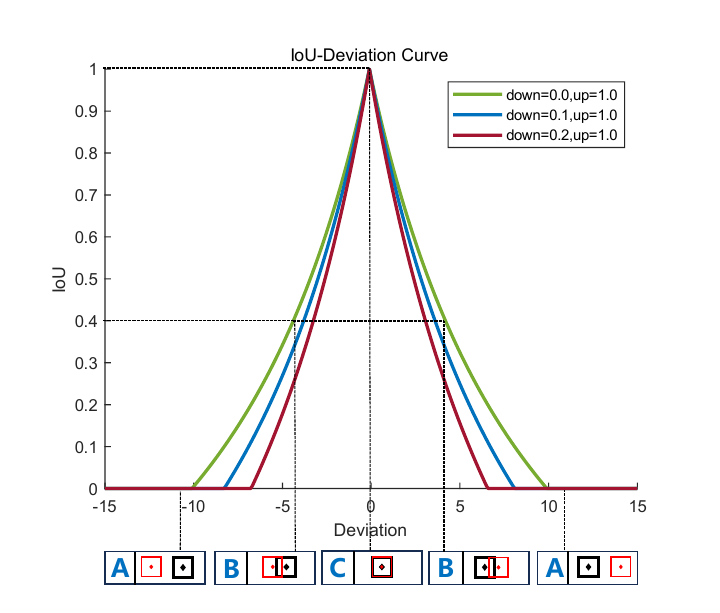}} 
	\caption{As shown in the figure:  (a) and (b) represent the linear interval mapping curves for difficult and simple samples, respectively.}
	\label{fig_3}
\end{figure*}
The problem of imbalanced samples exists in various object detection tasks, and can be divided into difficult samples and simple samples based on the difficulty of object detection. From the perspective of target scale analysis, general detection targets can be regarded as simple samples, while extremely small targets can be regarded as difficult samples due to their difficulty in precise positioning. For detection tasks dominated by simple samples, focusing on simple samples in the bounding box regression process will help improve detection performance. For detection tasks with a high proportion of difficult samples, in contrast, it is necessary to focus on the bounding box regression of difficult samples.
\subsection{Focaler-IoU}
For the purpose of focusing on different regression samples in being able to focus on different detection tasks, we use a linear interval mapping approach to reconstruct the IoU loss, which allows for improved edge regression. The formula is as follows:
\begin{equation}
	IoU^{focaler} = \begin{cases} 
		0, & IoU < d \\
		\frac{IoU - d}{u - d}, & d \ll IoU \ll u  \\
		1, & IoU>u \\
	\end{cases}
\end{equation}
Where $IoU^{focaler}$ is the reconstructed Focaler-IoU, IoU is the original IoU value, and $[d, u] \in [0, 1]$.
By adjusting the values of d and u, we can make $IoU^{focaler}$ focus on different regression samples.
Its loss is defined below:
\begin{equation}
L_{\text{Focaler-IoU}} = 1 - IoU^{focaler}
\end{equation}
Applying Focaler-IoU loss to the existing IoU based bounding box regression loss function, $L_{Focaler-GIoU}$, $L_{Focaler-DIoU}$, $L_{Focaler-CIoU}$, $L_{Focaler-EIoU}$ and $L_{Focaler-SIoU}$ are as follows:
\begin{equation}
	L_{Focaler-GIoU}=L_{GIoU}+IoU-IoU^{Focaler}
\end{equation}
\begin{equation}
	L_{Focaler-DIoU}=L_{DIoU}+IoU-IoU^{Focaler}
\end{equation}
\begin{equation}
	L_{Focaler-CIoU}=L_{CIoU}+IoU-IoU^{Focaler}
\end{equation}
\begin{equation}
	L_{Focaler-EIoU}=L_{EIoU}+IoU-IoU^{Focaler}
\end{equation}
\begin{equation}
	L_{Focaler-SIoU}=L_{SIoU}+IoU-IoU^{Focaler}
\end{equation}
\section{Experiments}
\subsection{PASCAL VOC on YOLOv8}
The PASCAL VOC dataset is one of the most popular datasets in the field of object detection, in this article we use the train and val of VOC2007 and VOC2012 as the training set including 16551 images, and the test of VOC2007 as the test set containing 4952 images. In this experiment, we choose the state-of-the-art one-stage detector YOLOv8s and YOLOv7-tiny to perform comparison experiments on the VOC dataset, and SIoU is chosen as the comparison method for the experiments. The experimental results are shown in TABLE\ref{tab:mytable1}:
\begin{table}[h]
	\centering
		\begin{tabular}{ccc}
		\toprule 
		& $AP_{50}$ & $mAP_{50:95}$ \\
		\midrule 
		Yolov8+SIoU & 69.5 & 48.3 \\
		Yolov8+Focaler-SIoU & 69.8(+0.3) & 48.6(+0.3) \\
		\bottomrule 
	\end{tabular}
		\caption{The performance of SIoU and Focaler-SIoU on Yolov8.}
	\label{tab:mytable1}
\end{table}
\subsection{AI-TOD on YOLOv5}
AI-TOD is a remote sensing image dataset, which is different from general datasets in that it contains a significant amount of tiny targets, and the average size of the targets is only 12.8 pixels. In this experiment, YOLOv5s is chosen as the detector, and the comparison method is SIoU. The experimental results are shown in TABLE\ref{tab:mytable3}:
\begin{table}[h]
	\centering
	\begin{tabular}{ccc}
		\toprule 
		& $AP_{50}$ & $mAP_{50:95}$ \\
		\midrule 
		Yolov5+SIoU & 42.7 & 18.1 \\
		Yolov5+Focaler-SIoU & 44.6(+1.9) & 18.6(+0.5) \\
		\bottomrule 
	\end{tabular}
	\caption{The performance of SIoU and Focaler-SIoU on Yolov5.}
	\label{tab:mytable3}
\end{table}
\section{Conclusion}
\par In this article, we analyzed the impact of the distribution of difficult and easy samples on object detection. When difficult samples dominate, it is necessary to focus on difficult samples to improve detection performance. When the proportion of simple samples is relatively large, the opposite is true. Next, we propose the Focaler-IoU method, which reconstructs the original IoU loss through linear interval mapping to achieve the goal of focusing on difficult and easy samples. Finally, comparative experiments have proven that the proposed method can effectively improve detection performance.

\end{document}